\documentclass{article} 
\usepackage{iclr2016_conference,times}
\usepackage{hyperref}
\usepackage{url}
\usepackage{amsmath,epsfig}
\usepackage{amsthm}
\usepackage{graphicx}
\usepackage{amssymb}
\usepackage{amsfonts}
\usepackage{tabulary}
\usepackage{multirow}
\usepackage{color}
\usepackage{colortbl}
\usepackage{wrapfig}
\usepackage{subfig}
\usepackage{epstopdf}

\definecolor{gry}{rgb}{0.92,0.92,0.92}

\title{Order-Embeddings of Images and Language}

\author{Ivan Vendrov, Ryan Kiros, Sanja Fidler, Raquel Urtasun \\
Department of Computer Science\\
University of Toronto\\
\texttt{\{vendrov,rkiros,fidler,urtasun\}@cs.toronto.edu} \\}
\iclrfinalcopy 

\newtheorem{definition}{Definition}
\begin{document}

\maketitle

\begin{abstract}

Hypernymy, textual entailment, and image captioning can be seen as special cases of a single visual-semantic hierarchy over words, sentences, and images. In this paper we advocate for explicitly modeling the partial order structure of this hierarchy. Towards this goal, we introduce a general method for learning ordered representations, and show how it can be applied to a variety of tasks involving images and language. We show that the resulting representations improve performance over current approaches for hypernym prediction and image-caption retrieval.


\end{abstract}

\vspace{-2mm}
\section{Introduction}
\vspace{-1mm}


Computer vision and natural language processing are becoming increasingly intertwined. Recent work in vision has moved beyond discriminating between a fixed set of object classes, to automatically generating open-ended lingual descriptions of images \citep{vinyals2015show}. Recent methods for natural language processing such as \citet{flickr30k} learn the semantics of language by grounding it in the visual world. Looking to the future, autonomous artificial agents will need to jointly model vision and language in order to parse the visual world and communicate with people.

\begin{figure}[h]
	\centering
	\includegraphics[width=0.5\textwidth]{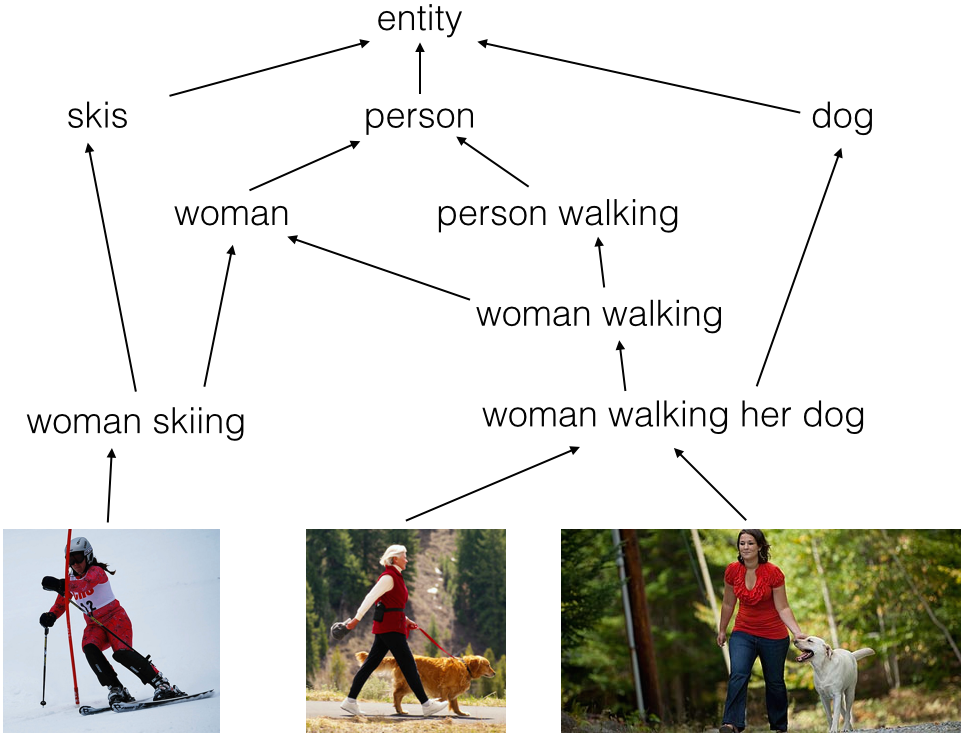}
	\caption{A slice of the visual-semantic hierarchy}
	\label{fig:sem}
\end{figure}

But what, precisely, is the relationship between images and the words or captions we use to describe them? It is akin to the hypernym relation between words, and textual entailment among phrases: captions are simply abstractions of images. In fact, all three relations can be seen as special cases of a partial order over images and language, illustrated in Figure \ref{fig:sem}, which we refer to as the \emph{visual-semantic hierarchy}. As a partial order, this relation is transitive: ``woman walking her dog", ``woman walking", ``person walking", ``person", and ``entity" are all valid abstractions of the rightmost image. Our goal in this work is to learn  representations that respect this partial order structure.

Most recent approaches to modeling the hypernym, entailment, and image-caption relations involve learning distributed representations or \emph{embeddings}. This is a very powerful and general approach which maps the objects of interest---words, phrases, images--- to points in a high-dimensional vector space.
One line of work, exemplified by \citet{chopra2005learning} and first applied to the caption-image relationship by \citet{socher2014grounded}, requires the mapping to be distance-preserving: semantically similar objects are mapped to points that are nearby in the embedding space. A symmetric distance measure such as Euclidean or cosine distance is typically used. Since the visual-semantic hierarchy is an \emph{anti}symmetric relation, we expect this approach to introduce systematic model error.

Other approaches do not have such explicit constraints, learning a more-or-less general binary relation between the objects of interest, e.g. \citet{bordes2011learning,socher2013reasoning,ma2015multimodal}. Notably, no existing approach directly imposes the transitivity and antisymmetry of the partial order, leaving the model to induce these properties from data. 

In contrast, we propose to exploit the partial order structure of the visual-semantic hierarchy by learning a mapping which is not distance-preserving but \emph{order}-preserving between the visual-semantic hierarchy and a partial order over the embedding space. We call embeddings learned in this way \emph{order-embeddings}. This idea can be integrated into existing relational learning methods simply by replacing their comparison operation with ours. By modifying existing methods in this way, we find that order-embeddings provide a marked improvement over the state-of-art for hypernymy prediction and caption-image retrieval, and near state-of-the-art performance for natural language inference.


This paper is structured as follows. We begin, in Section \ref{method}, by giving a unified mathematical treatment of our tasks, and describing the general approach of learning order-embeddings. In the next three sections we describe in detail the  tasks we tackle, how we apply the order-embeddings idea to each of them, and the results we obtain. The tasks are hypernym prediction (Section \ref{hypernyms}), caption-image retrieval (Section \ref{retrieval}), and textual entailment (Section \ref{entailment}).

In the supplementary material, we visualize novel vector regularities that emerge in our learned embeddings of images and language.

\section{Learning Order-Embeddings}
\label{method}
To unify our treatment of various tasks, we introduce the problem of \emph{partial order completion}.
In partial order completion, we are given a set of positive examples $P = \{(u,v)\}$ of ordered pairs drawn from a partially ordered set  $(X, \preceq_X)$, and a set of negative examples $N$ which we know to be unordered. Our goal is to predict whether an unseen pair $(u', v')$ is ordered. Note that  hypernym prediction, caption-image retrieval, and textual entailment are all special cases of this task, since they all involve classifying pairs of concepts in the (partially ordered) visual-semantic hierarchy.



We tackle this problem by  learning a mapping from $X$ into a partially ordered embedding space $(Y, \preceq_Y)$. The idea is to predict the ordering of an unseen pair in $X$ based on its ordering in the embedding space.  This is possible only if the mapping satisfies the following crucial property:
\begin{definition}
\label{eq:order}
A function $f : (X, \preceq_X) \to (Y, \preceq_Y)$ is an order-embedding if for all $u,v \in X$,
$$ u \preceq_X v \text{  if and only if  } f(u) \preceq_Y f(v) \label{eq:order-embedding} $$
\end{definition}

This definition implies that each combination of embedding space $Y$, order $\preceq_Y$, and order-embedding $f$ determines a unique completion of our data as a partial order $\preceq_X$. In the following, we first consider the choice of $Y$ and $\preceq_Y$, and then discuss how to find an appropriate $f$.

\subsection{The Reversed Product Order on $\mathbb{R}_{+}^N$}
The choice of $Y$ and $\preceq_Y$ is somewhat application-dependent. For the purpose of modeling the semantic hierarchy, our choices are narrowed by the following considerations.

Much of the expressive power of human language comes from \emph{abstraction} and \emph{composition}. 
For any two concepts, say ``dog" and ``cat", we can name a concept that is an abstraction of the two, such as ``mammal", as well as a concept that composes the two, such as ``dog chasing cat". 
So, in order to represent the visual-semantic hierarchy, we need to choose an order $\preceq_Y$ that is rich enough to embed these two relations.

We also restrict ourselves to orders  $\preceq_Y$ with a top element, which is above every other element in the order. In the visual-semantic hierarchy, this element represents the most general possible concept; practically, it provides an anchor for the embedding. 

Finally, we choose the embedding space $Y$ to be continuous in order to allow optimization with gradient-based methods.

A natural choice that satisfies all three properties is the reversed product order on $\mathbb{R}_{+}^N$, defined by the conjunction of total orders on each coordinate:
\begin{equation} 
x \preceq y \text{  if and only if  } \bigwedge_{i=1}^N x_i \geq y_i \label{eq:product-order} 
\end{equation}
for all vectors $x, y$ with nonnegative coordinates. Note the reversal of direction: smaller coordinates imply higher position in the partial order. The origin is then the top element of the order, representing the most general concept.

Instead of viewing our embeddings as single points $x \in \mathbb{R}_{+}^N$, we can also view them as sets $\{y :  x \preceq y\}$.  The meaning of a word is then the union of all concepts of which it is a hypernym, and the meaning of a sentence is the union of all sentences that entail it. The visual-semantic hierarchy can then be seen as a special case of the subset relation, a connection also used by \citet{flickr30k}.

\subsection{Penalizing Order Violations}
Having fixed the embedding space and order, we now consider the problem of finding an order-embedding into this space. 
In practice, the order embedding condition (Definition \ref{eq:order-embedding}) is too restrictive to impose as a hard constraint. Instead, we aim to find an \emph{approximate} order-embedding: a mapping which violates the order-embedding condition, imposed as a soft constraint, as little as possible.


More precisely, we define a penalty that measures the degree to which a pair of points violates the product order. In particular, we define the penalty for an ordered pair $(x,y)$ of points in $ \mathbb{R}_{+}^N$ as
\begin{equation} E(x,y) = ||\max(0, y-x)||^2  \label{eq:error}   \end{equation}
Crucially, $E(x,y) = 0 \iff x \preceq y $ according to the reversed product order; if the order is not satisfied, $E(x,y)$ is positive. This effectively imposes a strong prior on the space of relations, encouraging our learned relation to satisfy the partial order properties of transitivity and antisymmetry. This penalty is key to our method. Throughout the remainder of the paper, we will use it where previous work has used symmetric distances or learned comparison operators.

Recall that $P$ and $N$ are our positive and negative examples, respectively. Then, to learn an approximate order-embedding $f$, we could use a max-margin loss which encourages positive examples to have zero penalty, and negative examples to have penalty greater than a margin:
\begin{equation} \sum_{(u,v) \in P} E(f(u), f(v))   + \sum_{(u',v') \in N} \max\{0, \alpha - E(f(u'), f(v')) \}  \label{eq:pos} \end{equation}
In practice we are often not given negative examples, in which case this loss admits the trivial solution of mapping all objects to the same point. The best way of dealing with this problem depends on the application, so we will describe task-specific variations of the loss in the next several sections. 

\section{Hypernym Prediction}
\label{hypernyms}

To test the ability of our model to learn partial orders from incomplete data,  our first task is to predict withheld hypernym pairs in WordNet \citep{miller1995wordnet}. A {\it hypernym pair} is a pair of concepts  where the first concept is a specialization or an  instance of the second, e.g., (\emph{woman}, \emph{person}) or (\emph{New York}, \emph{city}).
 Our setup differs significantly from previous work in that we use \emph{only} the WordNet hierarchy as training data. The most similar evaluation has been that of \citet{baroni2012entailment}, who use external linguistic data in the form of distributional semantic vectors. \citet{bordes2011learning} and \citet{socher2013reasoning} also evaluate on the WordNet hierarchy, but they use other relations in WordNet as training data  (and external linguistic data, in Socher's case). 
 
 Additionally, the latter two consider only direct hypernyms, rather than the full, transitive hypernymy relation. But predicting the transitive hypernym relation is a better-defined problem because individual hypernym edges in WordNet vary dramatically in the degree of abstraction they require. For instance, \emph{(person, organism)} is a direct hypernym pair, but it takes eight hypernym edges to get from \emph{cat} to \emph{organism}.

\subsection{Loss Function}
\vspace{-0.5mm}

To apply  order-embeddings  to hypernymy, we follow the setup of \citet{socher2013reasoning} in learning an N-dimensional vector for each concept in WordNet, but  we replace their neural tensor network with our order-violation penalty defined in Eq. (\ref{eq:error}).
Just like them, we corrupt each hypernym pair by replacing one of the two concepts with a randomly chosen concept, and use these corrupted pairs as negative examples for both training and evaluation. We use their max-margin loss, which encourages the order-violation penalty to be zero for positive examples, and greater than a margin $\alpha$ for negative examples:
\begin{equation} \sum_{(u,v) \in WordNet} E(f(u), f(v))  + \max\{0, \alpha - E(f(u'), f(v')) \}  \label{eq:margin} \end{equation}
where $E$ is our order-violation penalty, and $(u', v')$ is a corrupted version of $(u, v)$. Since we learn an independent embedding for each concept, the mapping $f$ is simply a lookup table.

\vspace{-0.5mm}
\subsection{Dataset}
\vspace{-0.5mm}

The transitive closure of the WordNet hierarchy gives us $838073$ edges between $82192$ concepts in WordNet. Like \citet{bordes2011learning}, we randomly select $4000$ edges for the test split, and another $4000$ for the development set. Note that the majority of test set edges can be inferred simply by applying transitivity, giving us a strong baseline. 

\vspace{-0.5mm}
\subsection{Details of Training}
\vspace{-0.5mm}

We learn a 50-dimensional nonnegative vector for each concept in WordNet using the max-margin objective (\ref{eq:margin}) with margin $\alpha = 1$, sampling 500 true and 500 false hypernym pairs in each batch. We train for 30-50 epochs using the Adam optimizer \citep{adam} with learning rate $0.01$ and early stopping on the validation set. 
During evaluation, we find the optimal classification threshold on the validation set, then apply it to the test set.

\subsection{Results}

\begin{wrapfigure}[18]{r}{0.40\textwidth}
	\vspace{-4mm}
	\centering
	\includegraphics[width=0.40\textwidth]{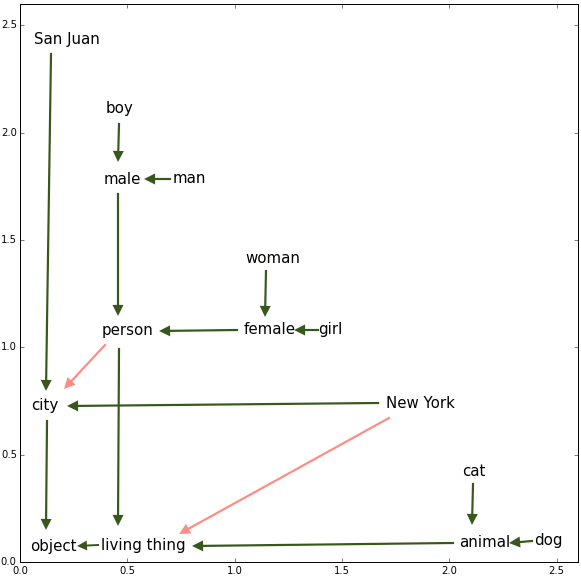}
	\vspace{-6mm}
	\caption{\small 2-dim order-embedding of a small subset of the WordNet hypernym relation. All the true hypernym pairs (green arrows) are correctly embedded, but two spurious pairs (pink arrows), are introduced. Only direct hypernyms are shown. }
	\label{fig:wordnet}
\end{wrapfigure}

Since our setup is novel, there are no published numbers to compare to. We therefore compare three variants of our model to two baselines, with results shown in Table \ref{wordnet}.

The \textbf{transitive closure} baseline involves no learning; it simply classifies hypernyms pairs as positive if they are in the transitive closure of the union of edges in the training and validation sets. The \textbf{word2gauss} baseline evaluates the approach of \citet{word2gauss} to represent words as Gaussian densities rather than points in the embedding space. This allows a natural representation of hierarchies using the KL divergence. We used 50-dimensional diagonal Gaussian embeddings, trained for 200 epochs on a max-margin objective with margin $7$, chosen by grid search\footnote{We used the code of \url{http://github.com/seomoz/word2gauss}}.

\textbf{order-embeddings (symmetric)} is our full model, but using symmetric cosine distance instead of our asymmetric penalty. \textbf{order-embeddings (bilinear)} replaces our penalty with the bilinear model used by \citet{socher2013reasoning}. \textbf{order-embeddings} is our full model.

Only our full model can do better than the transitive baseline, showing the value of exploiting partial order structure in contrast to using symmetric similarity or learning a general binary relation as most previous work and our bilinear baseline do.

The resulting 50-dimensional embeddings are difficult to visualize. To give some intuition for the structure being learned, Figure \ref{fig:wordnet} shows the results of a toy 2D experiment.

\begin{table}[t]
	\begin{center}
		\begin{tabular}{lc}
			\multicolumn{1}{c}{\bf Method}  &\multicolumn{1}{c}{\bf Accuracy (\%)}
			\\ \hline \\
			transitive closure & 88.2 \\
			word2gauss  &  86.6\\
			order-embeddings (symmetric) & 84.2 \\
			order-embeddings (bilinear) & 86.3 \\
			order-embeddings  & \textbf{90.6} \\
		\end{tabular}
	\end{center}
	\caption{Binary classification accuracy on 4000 withheld edges from WordNet.}
	\label{wordnet}
\end{table}

\section{Caption-Image Retrieval}
\label{retrieval}
The {\it caption-image retrieval} task has become a standard evaluation of joint models of vision and language  \citep{hodosh2013framing,LinCVPR14}. The task involves ranking a large dataset of images by relevance for a query caption (Image Retrieval), and ranking captions by relevance for a query image (Caption Retrieval). Given a set of aligned image-caption pairs as training data, the goal is then to learn a caption-image compatibility score $S(c,i)$ to be used at test time.

Many modern approaches model the caption-image relationship symmetrically, either by embedding into a common ``visual-semantic" space with inner-product similarity  \citep{socher2014grounded, kiros2014}, or by using Canonical Correlations Analysis between distributed representations of images and captions \citep{klein2015fisher}. While \citet{karpathydeep} and \citet{plummer2015flickr30k} model a finer-grained alignment between regions in the image and segments of the caption, the similarity they use is still symmetric.
An alternative is to learn an unconstrained binary relation, either with a neural language model conditioned on the image \citep{vinyals2015show,mao2015} or using a multimodal CNN \citep{ma2015multimodal}.

In contrast to  these lines of work, we propose to treat the caption-image pairs as a two-level partial order with captions above the images they describe, and let 
$$S(c,i) = - E(f_i(i), f_c(c)) $$
with $E$ our order-violation penalty defined in Eq (\ref{eq:error}), and $f_c, f_i$ are embedding functions from captions and images into $R_+^N$.

\subsection{Loss Function}
To facilitate comparison, we use the same pairwise ranking loss that \citet{socher2014grounded}, \citet{kiros2014} and \citet{karpathydeep} have used on this task---simply replacing their symmetric similarity measure with our asymmetric order-violation penalty. This loss function encourages $S(c,i)$ for ground truth caption-image pairs to be greater than that for all other pairs, by a margin:
\begin{equation} \sum_{(c, i)} \left( \sum_{c'} \max\{0, \alpha - S(c,i) + S(c', i)\}  + \sum_{i'} \max\{0, \alpha - S(c,i) + S(c, i')\} \right) \label{eq:contrastive} \end{equation}
where $(c,i)$ is a ground truth caption-image pair, $c'$ goes over captions that no describe $i$, and $i'$ goes over image not described by $c$.

\subsection{Image and Caption Embeddings}
To learn $f_c$ and $f_i$, we use the approach of \citet{kiros2014}  except, since we are embedding into $\mathbb{R}_{+}^N$, we constrain the embedding vectors to have nonnegative entries by taking their absolute value. Thus, to embed images, we use 
\begin{equation} 
f_i(i) = | W_i \cdot CNN(i) | 
\end{equation}
where $W_i$ is a learned $N\times 4096$ matrix, $N$ being the dimensionality of the embedding space. $CNN(i) $ is the same image feature used by \citet{klein2015fisher}: we rescale images to have smallest side $256$ pixels, we take $224\times 224$ crops from the corners, center, and their horizontal reflections, run the 10 crops through the 19-layer VGG network of \citet{vgg} (weights pre-trained on ImageNet and fixed during training), and average their \emph{fc7} features. 

To embed the captions, we use a recurrent neural net encoder with GRU activations \citep{gru}, so $f_c(c) = | GRU(c) |$, the absolute value of hidden state after processing the last word.

\subsection{Dataset}

We evaluate on the Microsoft COCO dataset \citep{coco}, which has over 120,000 images, each with at least five human-annotated captions per image. This is by far the largest dataset commonly used for caption-image retrieval. We use the data splits of \citet{karpathydeep} for training (113,287 images), validation (5000 images), and test (5000 images).

\subsection{Details of Training}

To train the model, we use the standard pairwise ranking objective from Eq.  (\ref{eq:contrastive}). We sample minibatches of 128 random image-caption pairs, and draw all contrastive terms from the minibatch, giving us 127 contrastive images for each caption and captions for each image. We train for 15-30 epochs using the Adam optimizer with learning rate $0.001$, and early stopping on the validation set.

We set the dimension of the embedding space and the GRU hidden state $N$ to $1024$, the dimension of the learned word embeddings to $300$,  and the margin $\alpha$ to $0.05$. All these hyperparameters, as well as the learning rate and batchsize, were selected using the validation set. For consistency with \citet{kiros2014} and to mitigate overfitting, we constrain the caption and image embeddings to have unit L2 norm. This constraint implies that no two points can be exactly ordered with zero order-violation penalty, but since we use a ranking loss, only the relative size of the penalties matters.

\subsection{Results}

\begin{table*}[t]
	\small
	\centering
	\begin{tabulary}{\linewidth}{L|CCCC|CCCC}
		& \multicolumn{4}{c}{Caption Retrieval} & \multicolumn{4}{c}{Image Retrieval} \\
		\textbf{Model} & \textbf{R@1} & \textbf{R@10} & \textbf{Med} \it{r}  & \textbf{Mean} \it{r}& \textbf{R@1} & \textbf{R@10} & \textbf{Med} \it{r}   & \textbf{Mean} \it{r}\\
		\hline
		& \multicolumn{8}{>{\columncolor{gry}}c}{}\\[-2.7mm]
		& \multicolumn{8}{>{\columncolor{gry}}c}{\textbf{1k Test Images}} \\[0.3mm]
		\cline{2-9}
		& & & & & & & \\[-2.5mm]
		MNLM \citep{kiros2014}  &  \underline{43.4} &  \underline{85.8} & 2 & * & 31.0 & 79.9 & 3 & *  \\
		$m$-RNN \citep{mao2015}  &41.0 & 83.5 & 2 & * & 29.0 & 77.0 & 3 & *  \\
		DVSA \citep{karpathydeep} & 38.4 & 80.5 & \textbf{1} & *  & 27.4  & 74.8 & 3 & * \\
		STV \citep{kiros2015skip} & 33.8  & 82.1 & 3 &  * & 25.9  & 74.6 & 4 & * \\
		FV \citep{klein2015fisher} & 39.4& 80.9 & 2 & 10.4 & 25.1 & 76.6 & 4 & 11.1 \\
		$m$-CNN \citep{ma2015multimodal} & 38.3  & 81.0 & 2 & * & 27.4 & 79.5 & 3 & * \\
		$m$-CNN$_{ENS}$ &  42.8 & 84.1 & 2 & * & 32.6 &  \underline{82.8} & 3 & *\\[0.6mm]
		\hline
		& & & & & & & \\[-2.5mm]
		order-embeddings (reversed)  & 11.2 & 44.0 & 14.2 & 86.6 & 12.3 & 53.5 & 9.0 & 30.1\\
		order-embeddings (1-crop)  & 41.4 &  84.2 & 2.0 & 8.7 &  \underline{33.5}  & 82.2 & \underline{2.6}& 10.0 \\
		order-embeddings (symm.) &  45.4  & 88.7 & 2.0 & 5.8  & 36.3 & 85.8 & \textbf{2.0} & 9.0 \\
		order-embeddings  & \textbf{46.7}  & \textbf{88.9} & 2.0 & \textbf{5.7} & \textbf{37.9} & \textbf{85.9} & \textbf{2.0} & \textbf{8.1}  \\[1.5mm]
		
		\hline
		& \multicolumn{8}{>{\columncolor{gry}}c}{}\\[-2.7mm]
		& \multicolumn{8}{>{\columncolor{gry}}c}{\textbf{5k Test Images}} \\[0.3mm]
		\cline{2-9}
		& & & & & & & \\[-2.5mm]
		DVSA & 11.8 & 45.4 & 12.2 & * & 8.9  & 36.3 & 19.5 & * \\
		FV & 17.3 & 50.2 & 10.0 & 46.4  & 10.8 & 40.1 & 17.0 & 49.3\\[0.6mm]
		\hline
		& & & & & & & \\[-2.5mm]
		order-embeddings (symm.) & 21.5 & 62.9 & 6.0 & 24.4 & 16.8  & 56.3 & 8.0 & 40.4 \\
		order-embeddings  & \textbf{23.3} & \textbf{65.0} & \textbf{5.0} & \textbf{24.4} & \textbf{18.0}  & \textbf{57.6} & \textbf{7.0} & \textbf{35.9} \\
		
	\end{tabulary}
	\caption{{ Results of caption-image retrieval evaluation on COCO. \textbf{R@K} is Recall@K, in \%. \textbf{Med} {\it r} is median rank. Metrics for our models on 1k test images are averages over five 1000-image splits of the 5000-image test set, as in \citep{klein2015fisher}. Best results overall are in bold; best results using 1-crop VGG features are underlined.}}
	\label{table:coco}
\end{table*}
Given a query caption or image, we sort all the images or captions of the test set in order of increasing penalty. We use standard ranking metrics for evaluation. We measure Recall@$K$, the percent of queries for which the GT term is one of the first $K$ retrieved; and median and mean rank, which are statistics over the position of the GT term in the retrieval order. 

Table \ref{table:coco} shows a comparison between all state-of-the-art  and some older methods\footnote{Note that the numbers for MNLM come not from the published paper but from the recently released code at \url{http://github.com/ryankiros/visual-semantic-embedding}.} along with our own; see \citet{ma2015multimodal} for a more complete listing. 

The best results overall are in bold, and the best results using 1-crop VGG image features are underlined. Note that the comparison is additionally complicated by the following:
\begin{itemize}
	\item $m$-CNN$_{ENS}$ is an ensemble of four different models, whereas the other entries are all single models.
	\item STV and FV use external text corpora to learn their language features, whereas the other methods learn them from scratch.
\end{itemize}

To facilitate the comparison and to evaluate the contributions of various components of our model, we evaluate four variations of order-embeddings: 

\textbf{order-embeddings} is our full model as described above.

\textbf{order-embeddings (reversed)} reverses the order of captions and image embeddings in our order-violation penalty---placing images above captions in the partial order learned by our model. This seemingly slight variation performs atrociously, confirming our prior that captions are much more abstract than images, and should be placed higher in the semantic hierarchy.

\textbf{order-embeddings (1-crop)} computes the image feature using just the center crop, instead of averaging over 10 crops.

\textbf{order-embeddings (symm.)} replaces our asymmetric penalty with the symmetric cosine distance, and allows embedding coordinates to be negative---essentially replicating MNLM, but with better image features. Here we find that a different margin ($\alpha = 0.2$) works best.

Between these four models, the only previous work whose results are incommensurable with ours is DVSA, since it uses the less discriminative CNN of \citet{alexnet} but 20 region features instead of a single whole-image feature.

Aside from this limitation, and if only single models are considered, order-embeddings significantly outperform the state-of-art approaches for image retrieval even when we control for image features. 

\subsection{Exploration}
Why would order-embeddings do well on such a shallow partial order? Why are they much more helpful for image retrieval than for caption retrieval? 

Intuitively, symmetric similarity should fail when an image has captions with very different levels of detail, because the captions are so dissimilar that it is impossible to map both their embeddings close to the same image embedding. Order-embeddings don't have this problem: the less detailed caption can be embedded very far away from the image while remaining above it in the partial order.

\begin{figure}[t!]
	\centering
	\begin{minipage}{0.292\linewidth}
		\vspace{0.8mm}
		\includegraphics[width=1\textwidth]{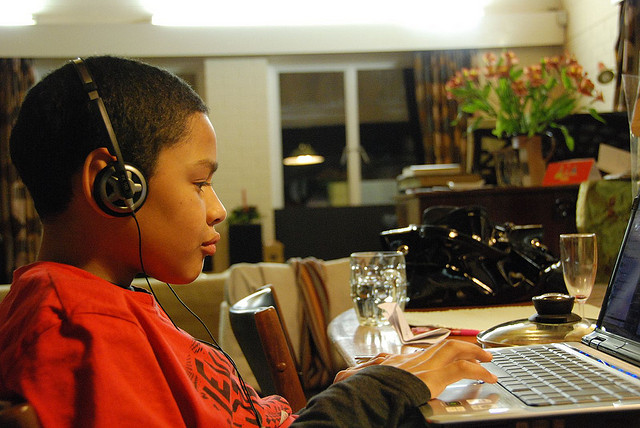}
	\end{minipage}
	\begin{minipage}{0.7\linewidth}
		\begin{small}
			\begin{tabular}{p{6.8cm}>{\centering\arraybackslash}p{0.7cm}>{\centering\arraybackslash}p{1.35cm}}
				{\bf Captions} & \multicolumn{2}{c}{\bf Image rank}\\
				& cosine & order-emb\\[1.8mm]
				a sitting area with furniture and flowers makes a backdrop for a boy with headphones sitting in the foreground at one of the chairs at a dining table that holds glasses and a handbag working at a laptop 
				& 4 & 8\\
				& & \\[-0.4mm]
				a kid is wearing headphone while on a laptop 
				& 286 & 24\\
			\end{tabular}
		\end{small}
		
	\end{minipage}
	\begin{minipage}{0.292\linewidth}
		\vspace{0.8mm}
		\includegraphics[width=\textwidth,trim=0 80 0 10,clip]{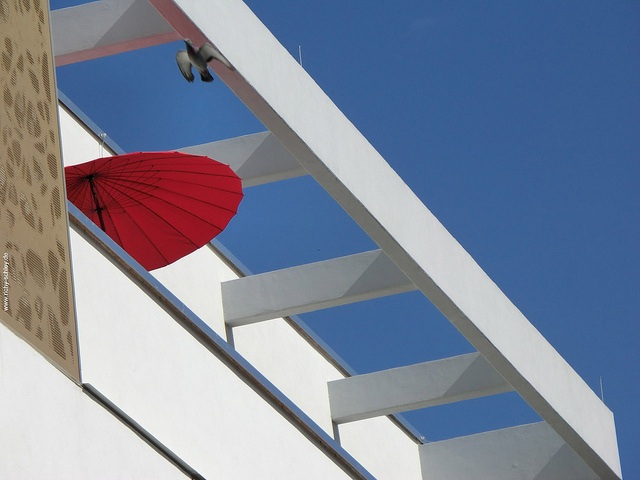}
	\end{minipage}
	\begin{minipage}{0.7\linewidth}
		\begin{small}
			\begin{tabular}{p{6.8cm}>{\centering\arraybackslash}p{0.7cm}>{\centering\arraybackslash}p{1.35cm}}
				view of top of a white building with tan speckled area an uncovered awning with a pigeon in fight below and a red umbrella behind balcony wall  
				& 3 & 5\\
				& & \\[-0.4mm]
				a pigeon flying near white beams of a building 
				& 91 & 6\\
			\end{tabular}
		\end{small}
	\end{minipage}		
	
	%

	\caption{Images with captions of very different lengths, and the rank of the GT image when using each caption as a query.}
	\label{fig:captions}
\end{figure}

To evaluate this intuition, we use caption length as a proxy for level of detail and select,  among pairs of co-referring captions in our validation set, the 100 pairs with the biggest length difference. For image retrieval with 1000 target images, the mean rank over captions in this set is $6.4$ for order-embeddings and $9.7$ for cosine similarity, a much bigger difference than over the entire dataset.  Some particularly dramatic examples of this are shown in Figure \ref{fig:captions}. Moreover, if we use the shorter caption as a query, and retrieve \emph{captions} in order of increasing error, the mean rank of the longer caption is $34.0$ for order-embeddings and $47.6$ for cosine similarity, showing that order-embeddings are able to capture the relatedness of co-referring captions with very different lengths.


This also explains why order-embeddings provide a much smaller improvement for caption retrieval than for image retrieval: all the caption retrieval metrics are based on the position of the \emph{first} ground truth caption in the retrieval order, so the embeddings need only learn to retrieve \emph{one} of each image's five captions well, which symmetric similarity is well suited for.

\section{Textual Entailment / Natural Language Inference}
\label{entailment}
Natural language inference can be seen as a generalization of hypernymy from words to sentences. For example, from ``woman walking her dog in a park" we can infer both ``woman walking her dog" and ``dog in a park", but not "old woman" or "black dog".  Given a pair of sentences, our task is to predict whether we can infer the second sentence (the hypothesis) from the first (the premise).

\vspace{-0.5mm}
\subsection{Loss Function}
To apply order-embeddings to this task, we again view it as partial order completion---we can infer a hypothesis from a premise exactly when the hypothesis is above the premise in the visual-semantic hierarchy.

Unlike our other tasks, for which we had to generate contrastive negatives,  datasets for natural language inference include labeled negative examples. So, we can simply use a max-margin loss:
\begin{equation} \sum_{(p,h)} E(f(p), f(h))  + \sum_{(p', h')} \max\{0, \alpha - E(f(p'), f(h')) \}  \label{eq:margin2} \end{equation}
where $(p, h)$ are positive and $(p', h')$ negative pairs of premise and hypothesis.
To embed sentences, we use the same GRU encoder as in the caption-image retrieval task.

\vspace{-0.5mm}
\subsection{Dataset}

To evaluate order-embeddings on the natural language inference task, we use the recently proposed SNLI corpus \citep{snli}, which contains 570,000 pairs of sentences, each labeled with ``entailment" if the inference is valid, ``contradiction" if the two sentences contradict, or ``neutral" if the inference is invalid but there is no contradiction.
Our method only allows us to discriminate between entailment and non-entailment, so we merge the ``contradiction" and ``neutral" classes together to serve as our negative examples. 

\vspace{-0.5mm}
\subsection{Implementation Details}
Just as for caption-image ranking, we set the dimensions of the embedding space and GRU hidden state to be $1024$, the dimension of the word embeddings to be $300$, and constrain the embeddings to have unit L2 norm. We train for 10 epochs with batches of $128$ sentence pairs. We use the Adam optimizer with learning rate $0.001$ and early stopping on the validation set. 
During evaluation, we find the optimal classification threshold on validation, then use the threshold to classify the test set.

\vspace{-1mm}
\subsection{Results}
The state-of-the-art method for 3-class classification on SNLI is that of \citet{rocktaschel2015reasoning}. Unfortunately, they do not compute 2-class accuracy, so we cannot compare to them directly.

As a bridge to facilitate comparison, we use a challenging baseline which can be evaluated on both the 2-class and 3-class problems. The baseline, referred to as \textbf{skip-thoughts},  involves a feedforward neural network on top of skip-thought vectors \citep{kiros2015skip}, a state-of-the-art semantic representation of sentences. Given pairs of sentence vectors $u$ and $v$, the input to the network is the concatenation of $u$, $v$ and the absolute difference $|u-v|$. We tuned the number of layers, layer dimensionality and dropout rates to optimize performance on the development set, using the Adam optimizer. Batch normalization \citep{ioffe2015batch} and PReLU units \citep{he2015delving} were used. Our best network used 2 hidden layers of 1000 units each, with dropout rate of 0.5 across both the input and hidden layers. We did not backpropagate through the skip-thought encoder.

We also evaluate against \textbf{EOP classifier}, a 2-class baseline introduced by \citep{snli}, and against a version of our model where our order-violation penalty is replaced with the symmetric cosine distance, \textbf{order-embeddings (symmetric)}.

\begin{table}[t]
	\begin{center}
		\begin{tabular}{lcc}
		
			\multicolumn{1}{c}{\bf Method}  &\multicolumn{1}{c}{\bf 2-class} &\multicolumn{1}{c}{\bf 3-class}
			\\ \hline \\
			Neural Attention \citep{rocktaschel2015reasoning} & * & \textbf{83.5}\\
			EOP classifier \citep{snli} & 75.0 & * \\
			skip-thoughts  & 87.7 & 81.5\\
			order-embeddings (symmetric) & 79.3 & *\\
			order-embeddings  & \textbf{88.6} & *\\
		\end{tabular}
	\end{center}
	\caption{Test accuracy (\%) on SNLI. }
		\label{table:snli}
	\vspace{-2mm}
\end{table}

The results for all models are shown in Table \ref{table:snli}. We see that order-embeddings outperform the skip-thought baseline despite not using external text corpora. While our method is almost certainly worse than the state-of-the-art method of \citet{rocktaschel2015reasoning}, which uses a word-by-word attention mechanism, it is also much simpler.

\vspace{-1mm}
\section{Conclusion and Future Work}
\vspace{-0.5mm}

We introduced a simple method to encode order into learned distributed representations, which allows us to explicitly model the partial order structure of the visual-semantic hierarchy. Our method can be easily integrated into existing relational learning methods, as we demonstrated  on three challenging tasks involving computer vision and natural language processing. On two of these tasks, hypernym prediction and caption-image retrieval, our methods outperform all previous work.

A promising direction of future work is to learn better classifiers on ImageNet \citep{imagenet}, which has over 21k image classes arranged by the WordNet hierarchy.  Previous approaches, including \citet{frome2013devise} and \citet{norouzi2013zero} have embedded words and images into a shared semantic space with symmetric similarity---which our experiments suggest to be a poor fit with the partial order structure of WordNet. We expect significant progress on ImageNet classification, and the related problems of one-shot and zero-shot learning,  to be possible using order-embeddings.

Going further, order-embeddings may enable learning the entire semantic hierarchy in a single model which jointly reasons about hypernymy, entailment, and the relationship between perception and language, unifying what have been until now almost independent lines of work.

%
%

\subsubsection*{Acknowledgments}
We thank Kaustav Kundu for many fruitful discussions throughout the development of this paper. The work was supported in part by an NSERC Graduate Scholarship. 

\bibliography{iclr2016_conference}

\begin{thebibliography}{32}
\providecommand{\natexlab}[1]{#1}
\providecommand{\url}[1]{\texttt{#1}}
\expandafter\ifx\csname urlstyle\endcsname\relax
  \providecommand{\doi}[1]{doi: #1}\else
  \providecommand{\doi}{doi: \begingroup \urlstyle{rm}\Url}\fi

\bibitem[Baroni et~al.(2012)Baroni, Bernardi, Do, and
  Shan]{baroni2012entailment}
Baroni, Marco, Bernardi, Raffaella, Do, Ngoc-Quynh, and Shan, Chung-chieh.
\newblock Entailment above the word level in distributional semantics.
\newblock In \emph{EACL}, 2012.

\bibitem[Bordes et~al.(2011)Bordes, Weston, Collobert, and
  Bengio]{bordes2011learning}
Bordes, Antoine, Weston, Jason, Collobert, Ronan, and Bengio, Yoshua.
\newblock Learning structured embeddings of knowledge bases.
\newblock In \emph{AAAI}, 2011.

\bibitem[Bowman et~al.(2015)Bowman, Angeli, Potts, and Manning]{snli}
Bowman, Samuel~R., Angeli, Gabor, Potts, Christopher, and Manning,
  Christopher~D.
\newblock A large annotated corpus for learning natural language inference.
\newblock In \emph{EMNLP}, 2015.

\bibitem[Cho et~al.(2014)Cho, Van~Merri{\"e}nboer, Gulcehre, Bahdanau,
  Bougares, Schwenk, and Bengio]{gru}
Cho, Kyunghyun, Van~Merri{\"e}nboer, Bart, Gulcehre, Caglar, Bahdanau, Dzmitry,
  Bougares, Fethi, Schwenk, Holger, and Bengio, Yoshua.
\newblock Learning phrase representations using rnn encoder-decoder for
  statistical machine translation.
\newblock In \emph{EMNLP}, 2014.

\bibitem[Chopra et~al.(2005)Chopra, Hadsell, and LeCun]{chopra2005learning}
Chopra, Sumit, Hadsell, Raia, and LeCun, Yann.
\newblock Learning a similarity metric discriminatively, with application to
  face verification.
\newblock In \emph{CVPR}, 2005.

\bibitem[Deng et~al.(2009)Deng, Dong, Socher, Li, Li, and Fei-Fei]{imagenet}
Deng, Jia, Dong, Wei, Socher, Richard, Li, Li-Jia, Li, Kai, and Fei-Fei, Li.
\newblock Imagenet: A large-scale hierarchical image database.
\newblock In \emph{CVPR}, 2009.

\bibitem[Frome et~al.(2013)Frome, Corrado, Shlens, Bengio, Dean, Mikolov,
  et~al.]{frome2013devise}
Frome, Andrea, Corrado, Greg~S, Shlens, Jon, Bengio, Samy, Dean, Jeff, Mikolov,
  Tomas, et~al.
\newblock Devise: A deep visual-semantic embedding model.
\newblock In \emph{NIPS}, 2013.

\bibitem[He et~al.(2015)He, Zhang, Ren, and Sun]{he2015delving}
He, Kaiming, Zhang, Xiangyu, Ren, Shaoqing, and Sun, Jian.
\newblock Delving deep into rectifiers: Surpassing human-level performance on
  imagenet classification.
\newblock \emph{ICCV}, 2015.

\bibitem[Hodosh et~al.(2013)Hodosh, Young, and Hockenmaier]{hodosh2013framing}
Hodosh, Micah, Young, Peter, and Hockenmaier, Julia.
\newblock Framing image description as a ranking task: Data, models and
  evaluation metrics.
\newblock \emph{JAIR}, 2013.

\bibitem[Ioffe \& Szegedy(2015)Ioffe and Szegedy]{ioffe2015batch}
Ioffe, Sergey and Szegedy, Christian.
\newblock Batch normalization: Accelerating deep network training by reducing
  internal covariate shift.
\newblock \emph{ICML}, 2015.

\bibitem[Karpathy \& Li(2015)Karpathy and Li]{karpathydeep}
Karpathy, Andrej and Li, Fei-Fei.
\newblock Deep visual-semantic alignments for generating image descriptions.
\newblock In \emph{CVPR}, 2015.

\bibitem[Kingma \& Ba(2015)Kingma and Ba]{adam}
Kingma, Diederik and Ba, Jimmy.
\newblock Adam: A method for stochastic optimization.
\newblock In \emph{ICLR}, 2015.

\bibitem[Kiros et~al.(2014)Kiros, Salakhutdinov, and Zemel]{kiros2014}
Kiros, Ryan, Salakhutdinov, Ruslan, and Zemel, Richard~S.
\newblock Unifying visual-semantic embeddings with multimodal neural language
  models.
\newblock \emph{arXiv preprint arXiv:1411.2539}, 2014.

\bibitem[Kiros et~al.(2015)Kiros, Zhu, Salakhutdinov, Zemel, Torralba, Urtasun,
  and Fidler]{kiros2015skip}
Kiros, Ryan, Zhu, Yukun, Salakhutdinov, Ruslan, Zemel, Richard~S, Torralba,
  Antonio, Urtasun, Raquel, and Fidler, Sanja.
\newblock Skip-thought vectors.
\newblock \emph{NIPS}, 2015.

\bibitem[Klein et~al.(2015)Klein, Lev, Sadeh, and Wolf]{klein2015fisher}
Klein, Benjamin, Lev, Guy, Sadeh, Gil, and Wolf, Lior.
\newblock Associating neural word embeddings with deep image representations
  using fisher vectors.
\newblock In \emph{CVPR}, 2015.

\bibitem[Krizhevsky et~al.(2012)Krizhevsky, Sutskever, and Hinton]{alexnet}
Krizhevsky, Alex, Sutskever, Ilya, and Hinton, Geoffrey~E.
\newblock Imagenet classification with deep convolutional neural networks.
\newblock In \emph{NIPS}, 2012.

\bibitem[Lin et~al.(2014{\natexlab{a}})Lin, Fidler, Kong, and
  Urtasun]{LinCVPR14}
Lin, Dahua, Fidler, Sanja, Kong, Chen, and Urtasun, Raquel.
\newblock Visual semantic search: Retrieving videos via complex textual
  queries.
\newblock In \emph{Proceedings of the IEEE Conference on Computer Vision and
  Pattern Recognition}, 2014{\natexlab{a}}.

\bibitem[Lin et~al.(2014{\natexlab{b}})Lin, Maire, Belongie, Hays, Perona,
  Ramanan, Doll{\'a}r, and Zitnick]{coco}
Lin, Tsung-Yi, Maire, Michael, Belongie, Serge, Hays, James, Perona, Pietro,
  Ramanan, Deva, Doll{\'a}r, Piotr, and Zitnick, C~Lawrence.
\newblock Microsoft coco: Common objects in context.
\newblock In \emph{ECCV}, 2014{\natexlab{b}}.

\bibitem[Ma et~al.(2015)Ma, Lu, Shang, and Li]{ma2015multimodal}
Ma, Lin, Lu, Zhengdong, Shang, Lifeng, and Li, Hang.
\newblock Multimodal convolutional neural networks for matching image and
  sentence.
\newblock \emph{ICCV}, 2015.

\bibitem[Mao et~al.(2015)Mao, Xu, Yang, Wang, and Yuille]{mao2015}
Mao, Junhua, Xu, Wei, Yang, Yi, Wang, Jiang, and Yuille, Alan.
\newblock Deep captioning with multimodal recurrent neural networks (m-rnn).
\newblock In \emph{ICLR}, 2015.

\bibitem[Mikolov et~al.(2013)Mikolov, Yih, and Zweig]{mikolov2013linguistic}
Mikolov, Tomas, Yih, Wen-tau, and Zweig, Geoffrey.
\newblock Linguistic regularities in continuous space word representations.
\newblock In \emph{HLT-NAACL}, pp.\  746--751, 2013.

\bibitem[Miller(1995)]{miller1995wordnet}
Miller, George~A.
\newblock Wordnet: a lexical database for english.
\newblock \emph{Communications of the ACM}, 1995.

\bibitem[Norouzi et~al.(2014)Norouzi, Mikolov, Bengio, Singer, Shlens, Frome,
  Corrado, and Dean]{norouzi2013zero}
Norouzi, Mohammad, Mikolov, Tomas, Bengio, Samy, Singer, Yoram, Shlens,
  Jonathon, Frome, Andrea, Corrado, Greg~S, and Dean, Jeffrey.
\newblock Zero-shot learning by convex combination of semantic embeddings.
\newblock In \emph{ICLR}, 2014.

\bibitem[Plummer et~al.(2015)Plummer, Wang, Cervantes, Caicedo, Hockenmaier,
  and Lazebnik]{plummer2015flickr30k}
Plummer, Bryan, Wang, Liwei, Cervantes, Chris, Caicedo, Juan, Hockenmaier,
  Julia, and Lazebnik, Svetlana.
\newblock Flickr30k entities: Collecting region-to-phrase correspondences for
  richer image-to-sentence models.
\newblock \emph{arXiv preprint arXiv:1505.04870}, 2015.

\bibitem[Rockt{\"a}schel et~al.(2015)Rockt{\"a}schel, Grefenstette, Hermann,
  Ko{\v{c}}isk{\`y}, and Blunsom]{rocktaschel2015reasoning}
Rockt{\"a}schel, Tim, Grefenstette, Edward, Hermann, Karl~Moritz,
  Ko{\v{c}}isk{\`y}, Tom{\'a}{\v{s}}, and Blunsom, Phil.
\newblock Reasoning about entailment with neural attention.
\newblock \emph{arXiv preprint arXiv:1509.06664}, 2015.

\bibitem[Simonyan \& Zisserman(2015)Simonyan and Zisserman]{vgg}
Simonyan, K. and Zisserman, A.
\newblock Very deep convolutional networks for large-scale image recognition.
\newblock In \emph{ICLR}, 2015.

\bibitem[Socher et~al.(2013)Socher, Chen, Manning, and Ng]{socher2013reasoning}
Socher, Richard, Chen, Danqi, Manning, Christopher~D, and Ng, Andrew.
\newblock Reasoning with neural tensor networks for knowledge base completion.
\newblock In \emph{NIPS}, 2013.

\bibitem[Socher et~al.(2014)Socher, Karpathy, Le, Manning, and
  Ng]{socher2014grounded}
Socher, Richard, Karpathy, Andrej, Le, Quoc~V, Manning, Christopher~D, and Ng,
  Andrew~Y.
\newblock Grounded compositional semantics for finding and describing images
  with sentences.
\newblock \emph{TACL}, 2014.

\bibitem[Van~der Maaten \& Hinton(2008)Van~der Maaten and Hinton]{tsne}
Van~der Maaten, Laurens and Hinton, Geoffrey.
\newblock Visualizing data using t-sne.
\newblock \emph{Journal of Machine Learning Research}, 9\penalty0
  (2579-2605):\penalty0 85, 2008.

\bibitem[Vilnis \& McCallum(2015)Vilnis and McCallum]{word2gauss}
Vilnis, Luke and McCallum, Andrew.
\newblock Word representations via gaussian embedding.
\newblock In \emph{ICLR}, 2015.

\bibitem[Vinyals et~al.(2015)Vinyals, Toshev, Bengio, and
  Erhan]{vinyals2015show}
Vinyals, Oriol, Toshev, Alexander, Bengio, Samy, and Erhan, Dumitru.
\newblock Show and tell: A neural image caption generator.
\newblock In \emph{CVPR}, 2015.

\bibitem[Young et~al.(2014)Young, Lai, Hodosh, and Hockenmaier]{flickr30k}
Young, Peter, Lai, Alice, Hodosh, Micah, and Hockenmaier, Julia.
\newblock From image descriptions to visual denotations: New similarity metrics
  for semantic inference over event descriptions.
\newblock \emph{TACL}, 2014.

\end{thebibliography}
\bibliographystyle{iclr2016_conference}

\clearpage
\newpage
\section{Supplementary Material}
\begin{figure}[h]
	\centering
	\includegraphics[width=\textwidth]{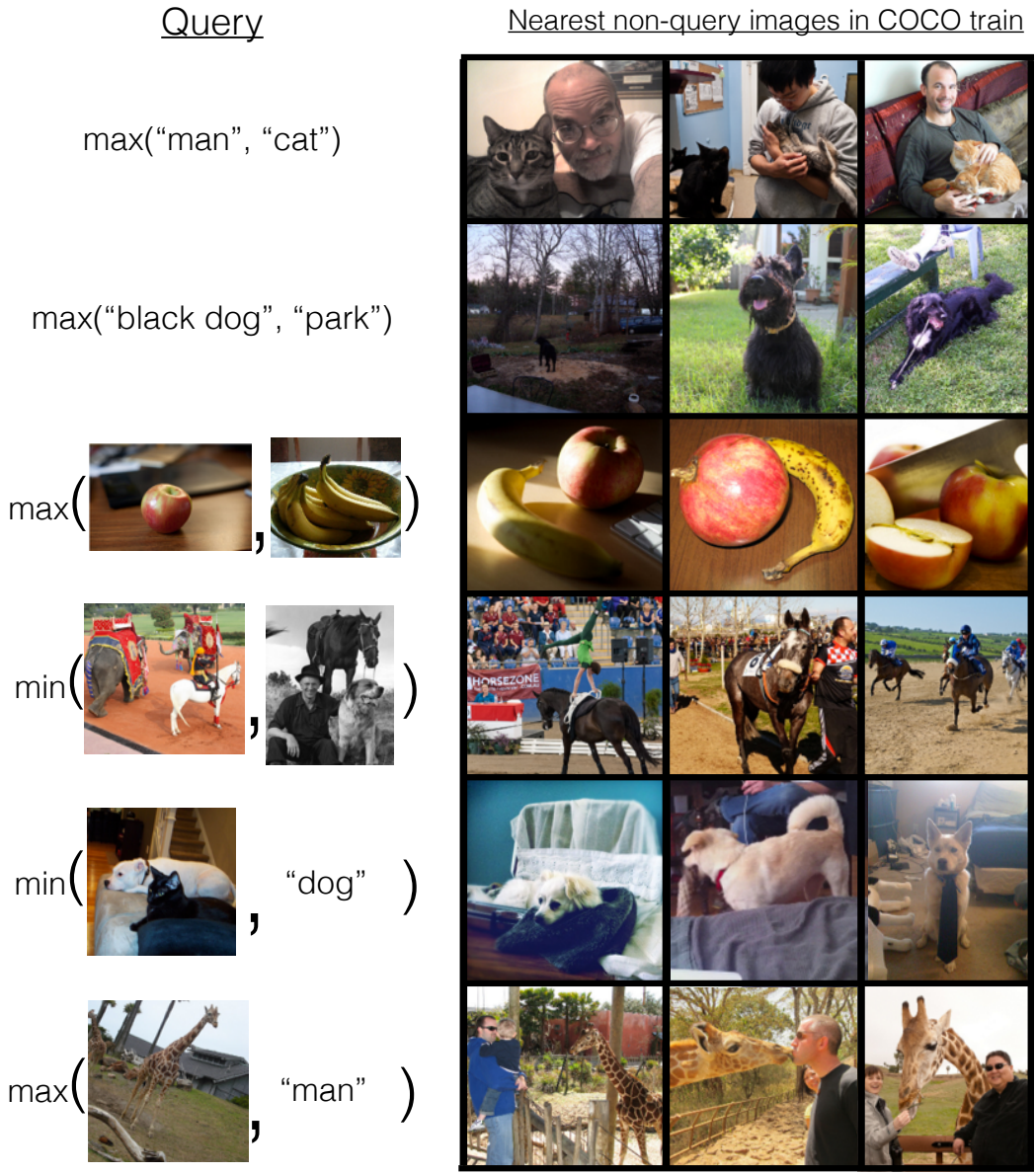}
	\caption{Multimodal regularities found with embeddings learned for the caption-image retrieval task. Note that some images have been slightly cropped for easier viewing, but no relevant objects have been removed. }
	\label{fig:regularities}
\end{figure}

\citet{mikolov2013linguistic} showed that word representations learned using word2vec exhibit semantic regularities, such as $king - man + woman \sim queen$. \citet{kiros2014} showed that  similar regularities hold for joint image-language models. We find that order-embeddings exhibit a novel form of regularity, shown in Figure \ref{fig:regularities}. The elementwise $\max$ and $\min$ operations in the embedding space roughly correspond to composition and abstraction, respectively.

\end{document}